\documentclass[10pt,twocolumn,letterpaper]{article}

\usepackage{cvpr}                % To produce the CAMERA-READY version
\usepackage{url}
\usepackage{xcolor}
\usepackage[normalem]{ulem}
\usepackage{soul}  

\newcommand{\pname}[1]{{CLOC}}
\newcommand{\loss}[1]{{MMNP}}
\newcommand{\losseq}[1]{{MM}}

\usepackage{graphicx}
\usepackage{multirow}
\usepackage{booktabs}
\usepackage{capt-of}
\usepackage{colortbl}
\usepackage{hhline}
\usepackage[accsupp]{axessibility}

\definecolor{cvprblue}{rgb}{0.21,0.49,0.74}
\usepackage[pagebackref,breaklinks,colorlinks,allcolors=cvprblue]{hyperref}

\def\paperID{13939} % *** Enter the Paper ID here
\def\confName{CVPR}
\def\confYear{2025}

\title{\pname{}: Contrastive Learning for Ordinal Classification with Multi-Margin N-pair Loss}

\author{  
    Dileepa Pitawela\textsuperscript{$\blacklozenge$} \\
  {\tt\small dileepa.pitawela@adelaide.edu.au}
\and
    Gustavo Carneiro\textsuperscript{$\blacktriangle$}\\
  {\tt\small g.carneiro@surrey.ac.uk}
\and
    Hsiang-Ting Chen\textsuperscript{$\blacklozenge$}\\
  {\tt\small tim.chen@adelaide.edu.au}
\and
    \textsuperscript{$\blacklozenge$}University of Adelaide, Australia $\:$
    \textsuperscript{$\blacktriangle$} CVSSP, University of Surrey, UK
}

\begin{document}

\maketitle

\begin{abstract}
In ordinal classification, misclassifying neighboring ranks is common, yet the consequences of these errors are not the same.
For example, misclassifying benign tumor categories is less consequential, compared to an error at the pre-cancerous to cancerous threshold, which could profoundly influence treatment choices. 
Despite this, existing ordinal classification methods do not account for the varying importance of these margins, treating all neighboring classes as equally significant. To address this limitation, we propose \pname{}, a new margin-based contrastive learning method for ordinal classification that learns an ordered representation based on the optimization of multiple margins with a novel multi-margin n-pair loss (\loss{}).
\pname{} enables flexible decision boundaries across key adjacent categories, facilitating smooth transitions between classes and reducing the risk of overfitting to biases present in the training data.
We provide empirical discussion regarding the properties of \loss{} and show experimental results on five real-world image datasets (Adience, Historical Colour Image Dating, Knee Osteoarthritis, Indian Diabetic Retinopathy Image, and Breast Carcinoma Subtyping) and one synthetic dataset simulating clinical decision bias.
Our results demonstrate that \pname{} outperforms existing ordinal classification methods and show the interpretability and controllability of \pname{} in learning meaningful, ordered representations that align with clinical and practical needs.
\let\thefootnote\relax\footnote{Code is available at \url{https://github.com/dpitawela/CLOC}}
\end{abstract}
%%%%%%%%%%%%%%%%%%%%%%%%%%%%%%%%%%%%%%%%%%%%
% Introduction

\begin{figure*}
\centering
\includegraphics[width=\textwidth]{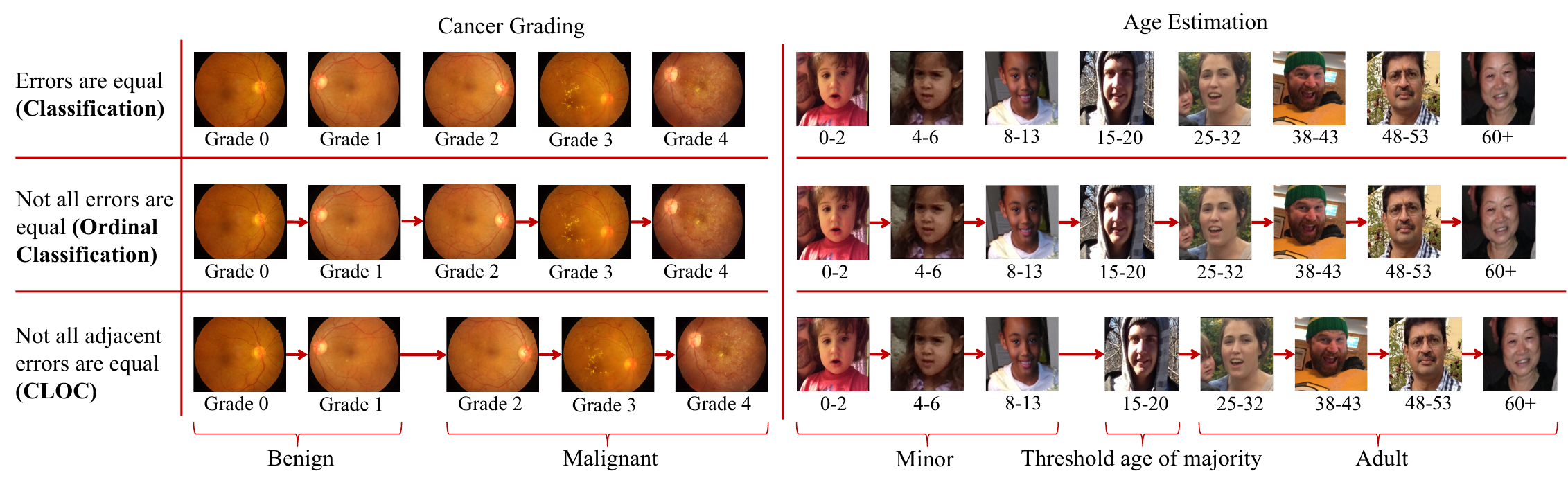}
\captionof{figure}{An illustration of the impact of classification errors in ordinal contexts. Standard classification treats all errors equally, while ordinal classification penalizes larger rank errors more heavily. \pname{} further accounts for the varying impact of adjacent errors, particularly across critical boundaries like benign versus malignant categories or minor vs threshold age of majority. This figure shows samples from IDRID (cancer grading) \citep{ds_idrid} and Adience (age estimation) \citep{ds_adience} datasets.
Longer arrow indicates a greater impact from adjacent errors.
}
\label{fig:teaser}
\end{figure*}

\section{Introduction}
Ordinal classification, the assignment of data to discrete, ordered categories, is essential across fields where ranking impacts key decisions. 
In healthcare, assessments such as TNM for cancer staging~\cite{Greene2002CancerStaging}, APACHE for intensive care \cite{Knaus1981apache}, and the Glasgow Coma Scale for brain injury~\cite{Teasdale2014ComaScale} are essential for guiding treatment and predicting outcomes. This approach is also widely employed in psychology (e.g., Likert scales), education (grading proficiency levels), and finance (credit scoring), where maintaining ordinal relationships is key to deriving meaningful insights that inform actionable decisions~\cite{gutierrez2015ordinal}.

Figure~\ref{fig:teaser} illustrates the impact of various types of classification errors in ordinal contexts.
Unlike standard classification problems, in ordinal classification or ranking problems, \textbf{not all label mistakes are equal}~\cite{lang_boosted,ranking_cnn,soft_labels_ord_reg}; misclassifications with larger rank differences carry more significant consequences. 
For example, they often represent greater discrepancies in disease severity or treatment needs. 
Moreover, \textbf{not all adjacent label mistakes are equally significant};
for instance, an error between two benign tumor categories may have less impact compared to a misclassification at the boundary between pre-cancerous and cancerous stages, where the shift in diagnosis could significantly alter treatment and management~\citep{birch2022clinical,Sox2024MedicalDecision}.
Similarly, a misclassification between minors and adults would have a larger impact than other adjacent ordinal categories~\cite{geng2013FacialAge}, as it could lead to more substantial legal or social implications. 

Existing methods have not been designed to address both of these challenges. \textit{Regression} attempts to model ordered categories, but its single continuous output often loses the ordinal structure and lacks clear boundaries between ranks~\cite{rnc, ordinalclip, clip_ordreg}. 
Additionally, its error penalties are less flexible, as typical regression or models, such as those minimizing mean squared error, do not inherently account for the ordinal relationships between classes nor allow modification.
\textit{Ordinal classification} methods~\cite{orcnn, ranking_cnn} are well-suited for tasks involving ordered categories, as they preserve the ranking structure and focus on maintaining the correct order, rather than merely predicting label distributions. However, most existing approaches focus on preserving order and minimizing label overlap in adjacent ranks, without considering the varying significance of adjacent label errors or examining the impact of margins between ranked labels.

We propose the new Contrastive Learning for Ordinal Classification (\pname{}) method that is based on an optimization that minimizes the novel Multi-Margin N-Pair (\loss{}) loss.
The new \loss{} loss is a central contribution of this paper that introduces multiple learnable margins that allow flexible decision boundaries between adjacent labels, while its cumulative property preserves class ordering. 
These multiple learnable margins improve generalization by preventing overfitting to specific biases or minor variations in the training data, while also enabling more robust transitions between classes than previous ordinal classifiers that rely on fixed, hard decision boundaries. Additionally, \pname{} enables nuanced ordinal classifications, making it particularly suited for challenging tasks like medical image analysis, where subtle distinctions between some categories are crucial.
Furthermore, these learnable margins also offer interpretability of the importance between ordinal categories and provide an option for manual adjustment, allowing flexible control over critical decision boundaries. 
To summarize, our key contributions are:
\begin{itemize}
    \item Novel Contrastive Learning for Ordinal Classification (\pname{}) method that is based on an optimization that minimizes the novel Multi-Margin N-Pair (\loss{}) loss, which provides learnable margins for flexible decision boundaries while preserving the ordinal structure.
    \item A novel method allowing users to control the model behavior through controlled margins, with experiments showing its effectiveness and robustness to biases.
    \item Enhanced interpretability via multi-margin learning, offering a new tool for interpretable AI and insights into class separability and decision boundaries in ordinal classification tasks.
\end{itemize}
We conducted experiments on five real-world image datasets (Adience, Historical Colour Image Dating, Knee Osteoarthritis, Indian Diabetic Retinopathy Image, and Breast Carcinoma Subtyping), as well as a synthetic dataset designed to simulate clinical decision biases. Our results show that \pname{} outperforms existing ordinal classification methods.
We also show empirical evidence of both CLOC's interpretability and controllability in learning meaningful, ordered representations that align closely with clinical and practical needs.

%%%%%%%%%%%%%%%%%%%%%%%%%%%%%%%%%%%%%%%%%%%%
% RELATED WORK
\section{Related Work}

\textbf{Ordinal Classification:}
Early ordinal classification approaches \cite{can2016regress, guo2008regress} relied on standard classification or regression, missing ordinal relationships and non-uniform label separations. 
More recent models like ORCNN \citep{orcnn} and RankingCNN \citep{ranking_cnn} use multiple binary classifiers to predict rank order, while CNNPOR \cite{constrained_net} introduced pairwise ordinal constraints to preserve structure. 
Methods like SORD \cite{soft_labels_ord_reg} and POEs \cite{poe} use soft labels and probabilistic embeddings, respectively, to better capture ordinal information.
Unlike these, which assume fixed class-pair error distributions that limit prioritization of critical boundaries, \pname{} enables class-pair-specific distributions and allows direct, meaningful adjustments over critical decision boundaries.

Order learning approaches such as pairwise comparators \citep{order_lim} and Deep Repulsive Clustering (DRC) \citep{order_leekim} compare instances directly to infer rank but are sensitive to reference quality. Moving Window Regression (MWR) \citep{mwr} refines ranks iteratively with global and local regressors, though it also depends on good reference selection.

Vision-language models like OrdinalCLIP \citep{ordinalclip}, Wang et al. \citep{lang_boosted}, and NumCLIP \citep{clip_ordreg} utilise CLIP for ordinal regression with rank-specific prompts and regularization, but these large models (190M+ parameters) are data-intensive. In contrast, our method efficiently captures ordinal relationships using a smaller model with contrastive learning, enabling scalable rank learning without heavy reliance on probabilistic distributions or large architectures.

\textbf{Representation learning:} 
Representation learning has been widely explored in classification, with contrastive learning becoming a leading approach, particularly in self-supervised contexts. 
The supervised variant, SupCon \cite{supcon}, has shown superior performance over traditional cross-entropy loss in tasks such as image recognition \citep{supcon}, long-tailed classification \citep{long_trailed_cls}, out-of-domain detection \citep{out_of_dom_det}, and anomaly detection \citep{anomaly_det}. 
While adaptations for ordinal tasks exist, such as in gaze estimation \citep{gaze_estimation} and medical imaging \citep{medical_cont1, medical_cont2}, these methods often fail to account for the ordinal nature of classes.

To address this, Xiao et al. \cite{xiao_et_al} proposed using labeled distances to capture semantic similarities and maintain local structure. Li et al. \cite{Li_et_al} introduced rank-proportional distance scaling, while Suarez et al. \cite{suarez_et_al} aligned embedding distances with rank differences. The work by \cite{gol} encoded order and metric relations using sample direction and distance. Further enhancements include Ordinal Log-Loss \citep{oll}, which uses distance-aware weighting, and RankSim \citep{ranksim}, which applies regularization for ordinal consistency. Rank-N-Contrast \cite{rnc} introduced regression-aware embeddings to align distances with continuous targets.

In contrast, our method inherently models ordinal relationships through contrastive learning without relying on predefined metrics or additional regularization, leading to more flexible and robust feature representations.

\textbf{Contrastive learning:}
CL focuses on maximizing similarity between positive pairs while separating negative ones, with temperature scaling and margin adjustments being critical parameters. \citet{Wang2021UnderstandingLoss} shows that temperature tuning helps balance feature alignment and tolerance for semantically similar samples, while \citet{Shah2022Max-Margin} introduces SVM-like margins in CL, though extending this to ordinal tasks is challenging. \citet{Rho2023UnderstandCL} explores single-margin adjustments for better generalization, but multi-margin strategies remain unexplored. Key CL frameworks like SimCLR \citep{simclr} and DINO \citep{dino} focus on unsupervised learning through augmentations and self-distillation, respectively. SupCon \citep{supcon} adapts CL for supervised tasks by clustering same-class samples, and Rank-N-Contrast (RNC) \citep{rnc} introduces a rank-based loss for ordinal representation. Our approach, \pname{}, builds on these advances with a novel multi-margin loss tailored for ordinal classification, showing superior performance and introducing better controllability and interpretability over decision boundaries across diverse datasets.

\section{Methodology}
Ordinal classification is a special case of multi-class classification where the labels
have a natural ordering or ranking among them.
Let the training dataset be
$\mathcal{D}=\{(x_i, y_i)\}_{i=1}^{N}$, where $x_i \in \mathcal{X}$ is a data sample and 
$y_i \in \mathcal{Y} = \{r_1,r_2,...,r_C\}$ is ground truth label from a $C$-class problem
that follow an ordered rank $r_1 \prec r_2 \prec ... \prec r_C$ 
with $C$ denoting the number of ranks, and $\prec$
indicating the ordering between different ranks. Similar to classification, ordinal classification seeks to predict
$\hat{y}_i \in \mathcal{Y}$ from $x_i$. 
Data are passed through an encoder $e_\phi(\cdot)$ to generate feature embeddings, $e_\phi: \mathcal{X} \to \mathcal{Z}; \mathcal{Z} \in \mathbb{R}^d$ 
% denoted as $z_i = e_\phi(x_i); \: z_i\in \mathcal{Z} \subset \mathbb{R}^d$ 
and followed by a classifier $c_\gamma(\cdot)$ to get output embeddings, 
$c_\gamma: \mathcal{Z} \to \mathcal{V}; \mathcal{V} \in \mathbb{R}^C$.
% $v_i = c_\psi(z_i); \: v_i \in \mathbb{R}^C$ 
The embeddings are subsequently converted to
$\hat{y} = \mathsf{OneHot}(\mathsf{softmax}(\mathbf{v}))$ where $\mathbf{v} \in \mathcal{V}$, $\mathsf{softmax}:\mathbb{R}^C \to \Delta^{C-1}$ and $\mathsf{OneHot}:\Delta^{C-1} \to \mathcal{Y}$, returning a one-hot label representing the class with the largest prediction. In this context, the terms `class', `label' and `rank' are used interchangeably to refer to the ordinal classification label.

\subsection{Contrastive Learning for Ordinal Classification (\pname{})}

\begin{figure}[t]
\centering
\includegraphics[width=\linewidth]{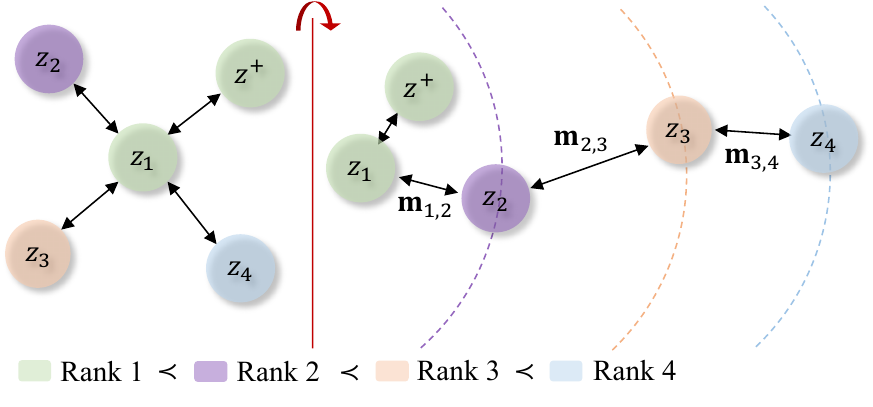}
\caption{Set of data samples from different classes/ranks (indicated by colors), with their rank order relative to the anchor $z_1$ and its positive sample $z^+$,
where $\text{rank } 1 \prec \text{rank } 2 \prec \text{rank } 3 \prec \text{rank } 4$.
The \loss{} loss pulls samples of the same rank closer to the anchor while pushing those with larger rank differences farther apart, resulting in wider margins in the representation space. Additionally, we can control specific margin values (e.g., by applying constraints) and use the learned margins to interpret class boundaries.}\label{fig:contrast_vis} 
\end{figure}
\pname{} addresses ordinal classification by learning a rank-aware representation through contrastive learning with multiple learnable classification margins. Unlike previous margin-based learning methods \cite{npair_loss}, which rely on a fixed, single margin, CLOC’s multiple learnable margins allow the model to better capture more nuanced separations between ranks, offering added flexibility for various applications. This approach not only provides finer control over specific margins but also enhances interpretability of the learned boundaries, addressing limitations in earlier fixed-margin methods.

Formally, in a $C$-class classification setting, we define $C-1$ pairs of consecutive ranks indexing the margins, represented by the set $\mathcal{H} = \{(r_1, r_2),(r_2, r_3), ..., (r_{C-1}, r_C)\}$. 
The overall optimization of \pname{} is based on minimizing the following objective function
\begin{equation}
\scalebox{0.90}{$
    \begin{split}
        \phi^*,\gamma^*,m^*_{\mathcal{H}} =&\arg\min_{\phi,\gamma,m_\mathcal{H}}  \frac{1}{N}\sum_{(x,y) \in \mathcal{D}}\ell_{CE}(y,\hat{y}) + \ell_{\losseq{}}(z,y, m_h)\\
        &\text{subject to } m_h > \rho, \forall h \in \mathcal{H}  
    \end{split}
    $}
\label{eq:overall_op}    
\end{equation}
where $m_{\mathcal{H}} = \{m_h\}_{h \in \mathcal{H}}$ with  $m_h \in \mathbb{R}$,
% $\ell(y,\hat{y},z,m_h) = \ell_{CE}(y,\hat{y}) + \ell_{\loss{}}(z,y, m_h)$,
$z=e_\phi(x)$, $\hat{y} = c_\gamma(z)$,
$\ell_{CE}(y,\hat{y})=-y^{\top}\log(\hat{y})$ which is cross-entropy, and
$\ell_{\losseq{}}(z,y)$ denotes the proposed \loss{} loss defined below in~\cref{sec:mmnp}.
Overall, \pname{} jointly optimizes $\phi,\gamma$ and $m_{\mathcal{H}}$ to learn a representation for ordinal classification that allows flexible decision boundaries between adjacent ranks and preserve the ordinal relationship across ranks. 

\subsection{Multi Margin N-Pair (\loss{}) Loss}
\label{sec:mmnp}
As shown in Fig.~\ref{fig:contrast_vis}, the \loss{} loss, denoted by $\ell_{MM}(\cdot)$ in Eq.\eqref{eq:overall_op}, follows a contrastive learning setup, where the
set of positive and negative samples for the anchor $(x,y) \in \mathcal{D}$, which produces the embedding $z= e_{\phi}(x)$, is defined respectively as 
$\mathcal{S}^{+}=\{(z_j, y_j) \mid (x_j,y_j) \in\mathcal{D}, z_j = e_{\phi}(x_j),  z_j \ne z, y_j = y\}$, and $\mathcal{S}^{-} = \{(z_k, y_k) \mid (x_k, y_k)\in\mathcal{D}, z_k=e_{\phi}(x_k),  y_k \neq y\}$.
In its simplest form, as an optimization over all pairs jointly, \loss{} is given by,
\begin{equation}
\label{eq:mmnp}
\scalebox{0.98}{$
\begin{split}
   &\ell_{\losseq{}}(z,y,m_h) = \\
   &\sum_{(z_j,y_j) \in \mathcal{S}^{+}} \sum_{(z_k,y_k) \in \mathcal{S}^{-}} \max ( 0, \: \mathsf{m}_{y, y_k} + \psi(z, z_k) 
   - \psi(z, z_j) ) 
\end{split}
$}
\end{equation}
where $\psi(\cdot,\cdot)$ represents cosine similarity function, and 
$\mathsf{m}_{y, y_k}=m_{y,y+1}+...+m_{y_{k-1},y_k}$ (inverse the index if $y_k \prec y$).
That is, $\mathsf{m}_{y, y_k}$ is the cumulative sum of the margins between the anchor's rank $y$ and the rank ${y_k}$ of a negative sample. In practice, instead of forming the positive and negative sets from the whole dataset, we form them within the mini-batches, as explained in Sec.~\ref{sec:training}.

\noindent\textbf{Properties of the \loss{} Loss:}
The \loss{} loss enhances contrastive learning by introducing cumulative multiple margins $\mathsf{m}_{y, y_k}$ in the embedding space.
Previous contrastive loss based methods used for ordinal classification repels all negative samples in \(\mathcal{S}^{-}\) from the anchor \( z \) equally, regardless of rank differences.
In contrast, \loss{} loss repels the negative samples from more distant ranks more strongly than those from closer ranks and thus preserve the ordinal relationship in the embedding space.
More specifically, minimizing the \loss{} loss imposes the condition $\psi(z, z_k) + \mathsf{m}_{y, y_k} \leq \psi(z, z_j)$. It means that given two negative samples $(z_l, y_l)$ and $(z_m, y_m)$ with ranks \( y \prec y_l \prec y_m \), the similarity between the anchor $z$ and \( z_m \) is constrained to be lower than that with \( z_l \) due to the cumulative margin \( \mathsf{m}_{y, y_m} = \mathsf{m}_{y, y_l} + \mathsf{m}_{y_l, y_m} \). 
Additionally, the actual margin value between samples belonging to two consecutive ranked classes represents the learned distance to robustly separate  samples from those two classes. 
Hence the learned margins can be used to interpret the difficulty in separating samples from consecutively ranked classes. 
Furthermore, note from Eq.~\eqref{eq:overall_op}, that it is possible to constrain $m_h$ to a minimum value $\rho$, which is helpful when one wants to minimize mistakes between critical classes. Discussions about controlability and intepretability are in sections~\ref{sec:controllability} and \ref{sec:interpretability}.

\subsection{Training}
\label{sec:training}

\begin{figure}[b]
\centering
\includegraphics[width=\linewidth]{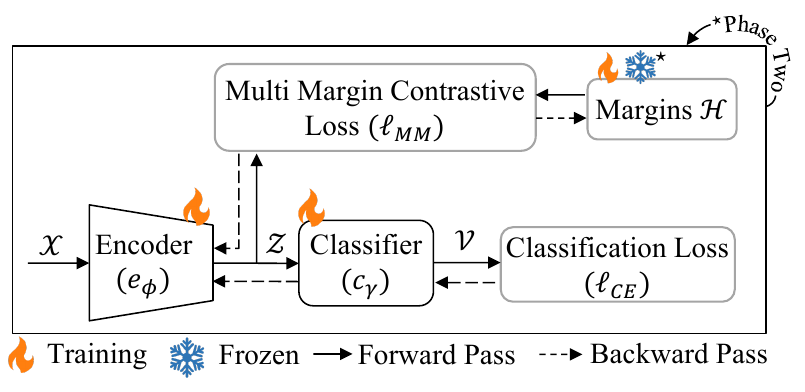}
\caption{Two phase training for \pname{}. In the first phase, the encoder, classifier and the margins are trained with the multi-margin contrastive loss and classification loss.
In the second phase, the margins are frozen allowing the training of the encoder and classifier with the losses.}
\label{fig:training}
\end{figure}

As shown in Figure~\ref{fig:training}, \pname{} employs a 2-phase training process ~\citep{anomaly_det_optim}.
In the first phase, the encoder, classifier and margins are jointly optimized to determine the margins that best separate classes and allow rank-aware representation, establishing a structured guide for representation learning.
In the second phase, these margins are frozen, allowing the model to refine the representation while adhering to the learned margins. 
The two-phase training strategy is employed to prevent the margins from collapsing to a trivial solution during the co-optimization of the representation and multiple margins, which we discuss below in this section, with additional details in Appendix \ref{sec:properties}.

Following the batch construction approach in \cite{npair_loss}, during both training phases, every mini-batch has, at least, samples from two distinct ranks, with a minimum of two samples taken from each rank. This setup ensures that each contrast includes at least one anchor, one positive sample, and two negative samples.

\subsubsection*{Phase One: Joint Learning of Margins and Representation}
\label{sec:phase_one}
In this phase, we run the optimization in~\eqref{eq:overall_op} to learn the margins $m^*_{\mathcal{H}}$, parameters of encoder $\phi^*$ and classifier $\gamma^*$ for producing the embeddings.
This phase seeks to optimize two objectives simultaneously: creating an ordered representation that is well-suited for classification, and learning margins that best separate ranks.
However, this dual objective introduces a potential risk of margin collapse, where all margins go to zero. Margins may converge toward zero to reduce the overall loss in Eq.~\eqref{eq:overall_op}, as setting margins to zero simplifies the optimization problem where the feature extractor is no longer incentivized to produce a nuanced ranked representation, potentially undermining the goal of learning structured, rank-separated embeddings.

To reduce the rate of margin collapse and enable training for a large number of epochs, we implement following ``precautionary'' measures: 1) the learnable margins are activated using the softplus function, that provides a smooth lower bound and approaches to zero gradually;
2) the margins are initialized with random values drawn from a uniform distribution in the interval $[0.5, 1.0)$, ensuring a moderate start avoiding excessively small or large initial margins; and
3) early stopping is applied to prevent collapse.

\subsubsection*{Phase Two: Margin-guided Representation Learning}
\label{sec:phase_two}
In the second phase, we optimize~\eqref{eq:overall_op} only with respect to $\phi$ and $\gamma$ with the learned margins being frozen.
This second phase ensures that the frozen margins that learned in phase one guide the feature extractor while continuing to improve the classification accuracy. 

The two-phase training allows integrating the precautionary mechanisms to phase one to address margin collapse.
In phase two, with margins fixed, the model exclusively focuses on refining the feature representation for better convergence to an optimal solution.

\section{Experiments}
We conducted three experiments evaluating the controllability,  interpretability, and overall performance of \pname{}.

\subsection{Datasets}
The \textbf{Adience} dataset \cite{ds_adience} serves as a benchmark for age prediction from facial images. It contains 26,580 colour images of 2,284 subjects, with each image categorized into one of eight age groups. 
We use the in-plane aligned version and five-folds originally released.
Each fold contains an average of 7,372 images, yeilding 7,372 test images and 29,490 training images per run. Per fold image count along with the average class distribution, is in appendix \ref{app:adience}. 

The \textbf{Historical Colour Image Dating (HID)} dataset \cite{ds_hid} is a benchmark for evaluating algorithms that predict the decade a colour image was captured. The dataset spans five decades, from the 1930s to the 1970s, and consists of 1,325 images, with 265 images for each decade. 
Similar to \cite{ds_hid, ord_reg1}, we run standard five-fold cross-validation by randomly selecting 50 images for testing, with the remaining 215 images for training from each decade.
Appendix \ref{app:hid} visualizes samples from this dataset. 

\begin{table*}[tb]
\centering
\arrayrulecolor{black}
\resizebox{0.77\linewidth}{!}{%
\begin{tabular}{l|ccccc|ccccc} 
\hline
\multicolumn{1}{c|}{\multirow{2}{*}{Method}} & \multicolumn{5}{c!{\color{black}\vrule}}{Accuracy$\uparrow$} & \multicolumn{5}{c}{Mean Absolute Error$\downarrow$} \\ 
\cline{2-11}
\multicolumn{1}{c|}{} & Adiance & HID & KOA & IDRID & BRACS & Adiance & HID & KOA & IDRID & BRACS \\ 
\hhline{>{\arrayrulecolor{black}}=>{\arrayrulecolor{black}}==========}
\multicolumn{11}{c}{Comparison with Ordinal Classification Methods} \\ 
\arrayrulecolor{black}\hline
ORCNN \citep{orcnn} & 0.5670 & 0.3870 & 0.4597 & 0.5833 & 0.4677 & 0.5400 & 0.9500 & 0.8335 & 0.5972 & 0.8732 \\
POE \citep{poe} & 0.6050 & 0.5468 & 0.6410 & 0.5730 & 0.5160 & 0.4700 & 0.7600 & 0.4440 & 0.6770 & 0.8280 \\
GOL \citep{gol} & 0.6250 & 0.5620 & 0.6439 & 0.6311 & 0.5754 & 0.4300 & \textbf{0.5500} & 0.4330 & 0.6890 & 0.7060 \\
MWR \citep{mwr} & 0.6260 & 0.5780 & 0.6380 & 0.6408 & 0.5649 & 0.4500 & 0.5800 & 0.5800 & 0.6752 & 0.8121 \\
RnC \textsuperscript{\textdagger} \citep{rnc} & 0.4679 & 0.5664 & 0.5886 & 0.5825 & 0.4632 & 0.8600 & 0.7288 & 0.5396 & 0.6152 & 0.8491 \\ 
\arrayrulecolor{black}\hline
\multicolumn{11}{c}{Comparison with Contrastive Learning Methods} \\ 
\arrayrulecolor{black}\hline
SimCLR \citep{simclr} & 0.3053 & 0.3988 & 0.5286 & 0.4563 & 0.3544 & 1.5303 & 1.0880 & 0.6964 & 1.0679 & 1.4859 \\
SupCon \citep{supcon} & 0.5774 & 0.4812 & 0.5955 & 0.5243 & 0.5667 & 0.5131 & 0.8040 & 0.4896 & 0.7476 & 0.7877 \\
DINO \citep{dino} & 0.4691 & 0.5004 & 0.5069 & 0.5437 & 0.5263 & 0.8230 & 0.8564 & 0.7555 & 0.7476 & 1.0719 \\
RnC \textsuperscript{\textdagger} \citep{rnc} & 0.4679 & 0.5664 & 0.5886 & 0.5825 & 0.4632 & 0.8600 & 0.7288 & 0.5396 & 0.6152 & 0.8491 \\ 
\arrayrulecolor{black}\hline
\pname{} (Ours) & \textbf{0.6302} & \textbf{0.6208} & \textbf{0.6673} & \textbf{0.7379} & \textbf{0.6035} & \textbf{0.4100} & 0.5534 & \textbf{0.4167} & \textbf{0.4078} & \textbf{0.7054} \\
\hline
\end{tabular}
}
\arrayrulecolor{black}
\captionof{table}{Comparison of test set accuracy and mean absolute error of \pname{} with contemporary ordinal classification methods (above) and contrastive learning methods (below) across five benchmark datasets. (\textsuperscript{\textdagger}RnC is both an ordinal classification and a contrastive method.)
}
\label{table:comparison_ord_n_cl}
\end{table*}
\begin{table*}[tb]
    \begin{minipage}[]{0.57\textwidth}
        \centering
        % \usepackage{graphicx}
% \usepackage{multirow}
% \usepackage{colortbl}
% \usepackage{hhline}

% \begin{table}
% \centering
% \arrayrulecolor{black}
\resizebox{0.93\linewidth}{!}{%
\begin{tabular}[t]{l|ccccccc|c} 
\hline
\multirow{2}{*}{\begin{tabular}[c]{@{}l@{}}\pname{}\\Margins\end{tabular}} & \multicolumn{7}{c|}{Error rate between class boundaries $\downarrow$ } & \multirow{2}{*}{\begin{tabular}[c]{@{}c@{}}Overall\\Acc.$\uparrow$\end{tabular}} \\ 
\cline{2-8}
 & \multicolumn{1}{c|}{C1 - C2} & \multicolumn{1}{c|}{C2 - C3} & \multicolumn{1}{c|}{C3 - C4} & \multicolumn{1}{c|}{C4 - C5} & \multicolumn{1}{c|}{C5 - C6} & \multicolumn{1}{c|}{C6 - C7} & C7 - C8 &  \\ 
\hhline{>{\arrayrulecolor{black}}=>{\arrayrulecolor{black}}=======>{\arrayrulecolor{black}}=}
\multicolumn{1}{l}{} & \multicolumn{7}{c}{IDRID} & \multicolumn{1}{l}{} \\ 
\arrayrulecolor{black}\hline
Learned  & 0.1282 & {\cellcolor[rgb]{0.91,0.91,0.91}}0.0270 & 0.0980 & 0.0937 & ~ & ~ & ~ & 0.7379 \\
Controlled  & 0.1025 & {\cellcolor[rgb]{0.91,0.91,0.91}}0.0000 & 0.1176 & 0.1562 &  &  &  & 0.6990 \\ 
\hline
\multicolumn{1}{l}{~} & \multicolumn{7}{c}{KOA} & \multicolumn{1}{l}{~} \\ 
\hline
Learned  & 0.2220 & {\cellcolor[rgb]{0.91,0.91,0.91}}0.1700 & 0.0985 & 0.0620 & ~ & ~ & ~ & 0.6673 \\
Controlled  & 0.2480 & {\cellcolor[rgb]{0.91,0.91,0.91}}0.0900 & 0.1160 & 0.0620 &  &  &  & 0.6431 \\ 
\hline
\multicolumn{1}{l}{~} & \multicolumn{7}{c}{HID} & \multicolumn{1}{l}{~} \\ 
\hline
Learned  & 0.0300 & 0.1300 & {\cellcolor[rgb]{0.91,0.91,0.91}}0.1500 & 0.2500 &  &  & ~ & 0.6280 \\
Controlled  & 0.0200 & 0.1700 & {\cellcolor[rgb]{0.91,0.91,0.91}}0.0800 & 0.2700 &  &  &  & 0.5640 \\ 
\hline
\multicolumn{1}{l}{~} & \multicolumn{7}{c}{Bracs} & \multicolumn{1}{l}{~} \\ 
\hline
Learned  & 0.1750 & 0.0931 & {\cellcolor[rgb]{0.91,0.91,0.91}}0.0545 & 0.0555 & {\cellcolor[rgb]{0.91,0.91,0.91}}0.1950 & 0.0602 & ~ & 0.6035 \\
Controlled  & 0.2560 & 0.0740 & {\cellcolor[rgb]{0.91,0.91,0.91}}0.0420 & 0.0490 & {\cellcolor[rgb]{0.91,0.91,0.91}}0.0850 & 0.0180 & ~ & 0.4281 \\ 
\hline
\multicolumn{1}{l}{~} & \multicolumn{7}{c}{Adience} & \multicolumn{1}{l}{~} \\ 
\hline
Learned  & 0.8410 & 0.0955 & {\cellcolor[rgb]{0.91,0.91,0.91}}0.1096 & 0.1899 & 0.2991 & 0.1522 & 0.2968 & 0.6134 \\
Controlled  & 0.8600 & 0.1210 & {\cellcolor[rgb]{0.91,0.91,0.91}}0.0640 & 0.2382 & 0.3247 & 0.1294 & 0.1718 & 0.5591 \\
\hline
\end{tabular}
}
\caption{Comparison of the overall accuracy and error rates for each pair of adjacent ranked classes between the original CLOC (Learned), where all margins are learned, and a modified version with fixed margins at critical decision boundaries (Controlled). Critical decision boundaries are highlighted in table~\ref{table:controllability}.}
% ,which learns all , and the training where the critical  (indicated by the highlighted cell in grey) is fixed (Controlled ). Critical boundaries are highlighted.}
% \arrayrulecolor{black}
% \end{table}
        \label{table:controllability}
    \end{minipage}%
    % \hspace{0.06cm}
    \hfill
    \begin{minipage}[*t]{0.41\textwidth}
        \centering
        \includegraphics[width=1\linewidth]{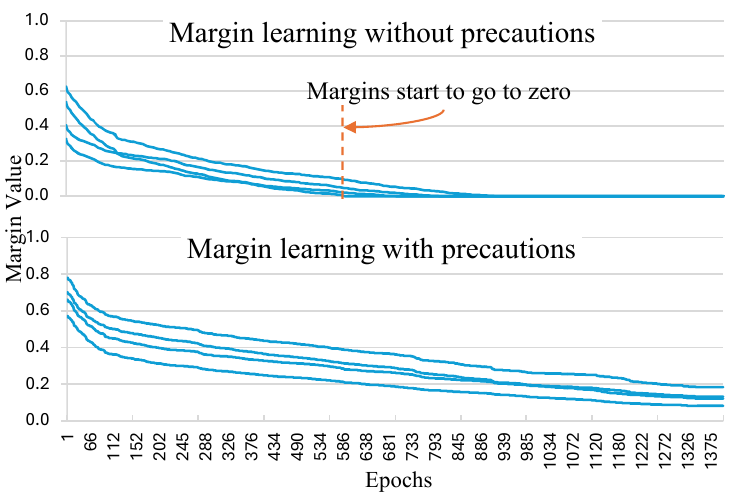}
        \captionof{figure}{Training phase one with and without margin collapse ``precaution'' measures. Notice that training continues for larger number of epochs without margin collapse when ``precaution'' measures are applied.}
        \label{fig:precautions}
    \end{minipage}
\end{table*}

The \textbf{Knee Osteoarthritis (KOA)} dataset \cite{ds_koa} aims to assess severity by the Kellgren-Lawrence (KL) grading system, where grade 0 represents a healthy knee and grade 4 indicates severe KOA. The dataset consists of 8,260 X-ray images, left and right joints combined. We train with 6,604 images and evaluate with 1,656 images. The class distribution of splits is in appendix \ref{app:koa}. 
The \textbf{Indian Diabetic Retinopathy Image Dataset (IDRiD)} \cite{ds_idrid} is designed to classify retinal fundus images into diabetic retinopathy grades defined by international clinical Diabetic Retinopathy (DR) scale. This scale identifies 5 grades, from 0 being no apparent DR to 4 being severe DR. The dataset contains 516 fundus images, split into 413 training images and 103 testing images, where each is assigned one of the 5 DR grades. The appendix \ref{app:idrid} details the class distribution of splits. 
The \textbf{BReAst Carcinoma Subtyping (BRACS)} dataset \cite{ds_bracs} facilitates classification of breast tumors from histological images. The grading system identifies 7 different subtypes of lesions ranging from grade 0 being normal to grade 6 being invasive carcinoma. The dataset includes 4,539 regions of interests from 189 patients split into 3,969 training and 570 testing samples. The class distribution of the dataset is in appendix \ref{app:bracs}.

\subsection{Experimental Setting}
\label{sec:ex_setting}
For Adience, the in-plane aligned version of the dataset is used, following the standard five-fold, subject-exclusive cross-validation protocol outlined in \cite{ds_adience, adience_param}. The image size is set to 224x224 pixels, with random horizontal and vertical flips, affine transformations, and color jittering applied as data augmentations.

For HID, as described in \cite{ds_hid, ord_reg1}, standard five-fold cross-validation is performed by randomly selecting 50 images for testing and using the remaining 215 images for training from each decade. Training involves a random crop of 224x224 pixels, along with random horizontal and vertical flips, affine and perspective augmentations, and the AutoAugment protocol from \cite{autoaug_cifar} with a probability of 0.01. For testing, a center crop of 224x224 pixels is used.

For KOA, a center crop of the image at 120x200 pixels is applied, followed by histogram equalization.
During training, random affine transformations, Gaussian blur, and perspective augmentations are used.

For both IDRID and BRACS, the image size is set to 224x224 pixels. During training, random horizontal and vertical flips, rotations, and Gaussian blur are applied. Additionally, random color jittering is used for IDRID, while random perspective augmentations are applied for BRACS.

For all experiments, an ImageNet pre-trained ResNet-50 \citep{resnet50} encoder is utilized, followed by two linear layers as the classifier.
Both training phases use Adam optimizer with a learning rate of 0.001, with $\rho = 0$ (as per eq.\ref{eq:overall_op}) and trains for 500 epochs.
Early stopping is applied in Phase One when training accuracy reaches 95\% and in Phase Two after 10 unimproved epochs. All experiments are performed on an RTX 4090 GPU.

We quantitatively assess performance by measuring accuracy and mean absolute error (MAE) on samples belonging to the test set of the datasets introduced above.

\subsection{Comparison with Related Methods}
Our experimental results, summarized in Table \ref{table:comparison_ord_n_cl}, compare \pname{} against both ordinal classification and contrastive learning baselines across five datasets. For ordinal classification, we include ORCNN \cite{orcnn}, POE \cite{poe}, MWR \cite{mwr}, GOL \cite{gol}, RnC \cite{rnc} and for contrastive learning methods, we cover SimCLR \citep{simclr}, DINO \citep{dino}, SupCon \citep{supcon}, and RNC \citep{rnc}.
Each contrastive learning method is evaluated using a ResNet-50 backbone pre-trained on ImageNet, with the full model re-trained on each respective training set and assessed on the corresponding test set.
The results show that \pname{} outperforms all other methods across every dataset.
In particular, notice that for some datasets, such as HID and IDRID, CLOC is substantially better than competing methods, with accuracy 0.6202 (compared to 0.5664 of the second best--RnC) on HID and 0.7379 (compared to 0.6408 of the second best--MWR) on IDRID.

In addition, we provide an experiment with large number of classes in Appendix \ref{sec:many_classes} and a computational time comparison with related methods in Appendix \ref{sec:runtime_comp}.

\subsection{Controllability and Robustness to Biases}
\label{sec:controllability}
As shown in Figure~\ref{fig:teaser}, there can be truly consequential classification boundaries in a dataset, such as between cancer and non-cancer classification in a diagnostic medical imaging dataset, or at the age of majority in a face dataset. 
As explained in Sec.~\ref{sec:mmnp}, our method allows the use of constraints in the learning of such critical margins, which is empirically demonstrated in this section.

\begin{figure*}[tb]
\centering
\includegraphics[width=0.95\textwidth]{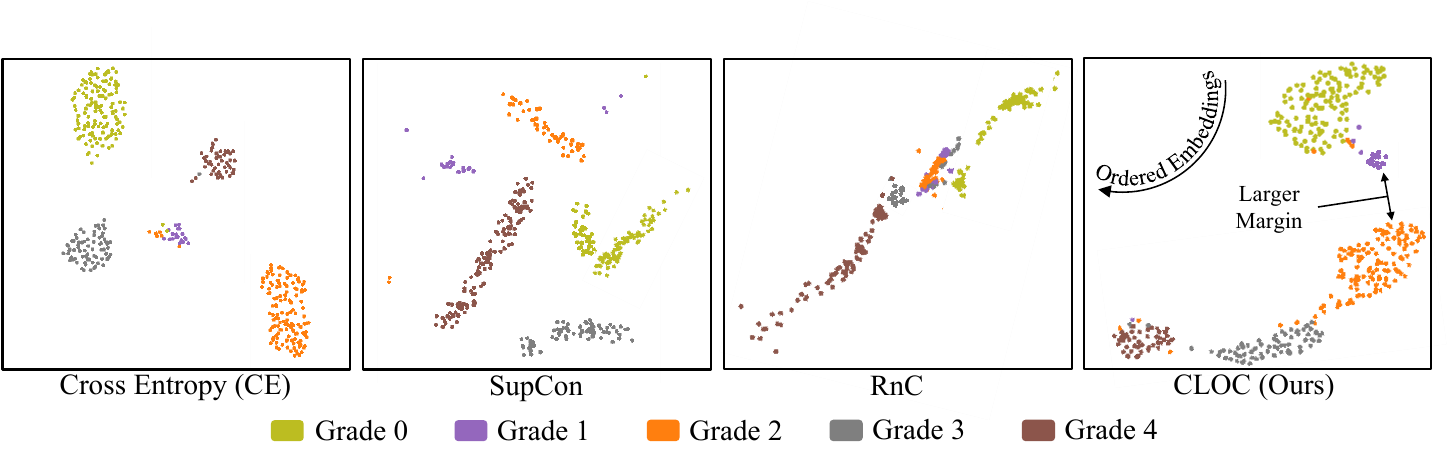}
\caption{UMAP visualizations of learned representations for the IDRID dataset that focuses on cancer grade classification. 
Notice that proposed \pname{}, effectively captures the inherent ordinal structure of cancer grades and maintains a larger margin at critical decision boundary, compared to other decision boundaries. Additional visualizations of GOL \cite{gol} and POE \cite{poe} are in Appendix \ref{sec:more_visuals}.}
\label{fig:emb_comp}
\end{figure*}

\noindent\textbf{Controlling the learning of critical margins:}
This experiment examines the effectiveness of manually fixing the margin to reduce errors between adjacent classes that intersect a key decision boundary.
For the IDRID and KOA datasets, the key decision boundaries that influence treatment choices lie between grades 1 and 2 \citep{idrid_distinct, koa_distinct}.
BRACS has two critical boundaries to separate benign grade 2 from atypical grade 3, and atypical grade 4 from malignant grade 5. 
For the Adience dataset, a practical critical decision boundary would be between age groups 8-13 and 15-20 to separate minors from threshold age of majority. 
For completeness, we assume a hypothetical critical boundary between the era of 50s and 60s in the HID dataset.

During the 2-phase optimization, we fix a larger value for the aforementioned critical decision margins and only optimize the remaining margins.
The rest of the experiment settings were identical to Section~\ref{sec:ex_setting}.

Table \ref{table:controllability} presents a comparison of overall accuracy and individual error rates between adjacent ranks for the learned margins approach (where all margins are optimized) and the controlled margin approach (where margin values are fixed at the critical decision boundary). Under the controlled margin condition, we observe a reduction in classification errors between classes near the critical boundary, albeit at the cost of lower overall accuracy.

\noindent\textbf{Robustness to Biases:}
This experiment simulates a scenario of \textit{diagnostic bias}. Studies in the medical domain have shown that practitioners have different subjective assessments~\cite{avery2024noninvasive,petashvili2024learning} and often lean towards positive diagnoses in high-stake cases~\cite{crowe2015Mental,roborgh2021risk}, a tendency explained by Error Management Theory (EMT) \cite{haselton2000EMT} and risk aversion. 
EMT suggests that when errors have unequal costs, people tend to err on the side of the less costly mistake; in medicine, a false negative (missed diagnosis) is typically viewed as more severe than a false positive. 
This bias, combined with doctors’ inherent risk aversion to potentially missing a critical diagnosis, drives them to favor caution, increasing the likelihood of positive diagnoses even at the expense of false positives.

The experiment simulates diagnostic bias in the IDRID training set at a critical decision boundary between grades 1 and 2 by randomly relabeling 60\% of the grade 1 samples as grade 2 and 30\% of the grade 2 samples as grade 1. 
We report both the results of using learnable margins, where the model learns all margins automatically, and the results of a manually set margin that enforces a fixed, larger margin between grades 1 and 2.
\cref{table:noise} presents a comparison with related methods on simulated biased training data. The results demonstrate that \pname{} with a fixed critical decision margin is significantly more robust to label noise compared to \pname{} with fully learnable margins. Compared to related methods, \pname{} achieves accuracy gains exceeding 7.9\% over the second best--GOL (\cref{table:noise}).

\begin{table*}[t]
\begin{minipage}[t]{0.3\linewidth}
    \centering
    \resizebox{0.81\linewidth}{!}{
    % \usepackage{graphicx}
% \usepackage{colortbl}

% \begin{table}[tb]
% \centering
% \arrayrulecolor{black}
% \resizebox{0.6\linewidth}{!}{%
% \begin{tabular}[t]{l|cc} 
% \hline
% Method & Accuracy$\uparrow$ & MAE$\downarrow$ \\ 
% \hline\hline
% SupCon \cite{supcon} & 0.4660 & 0.9029 \\
% RnC \cite{rnc} & 0.5437 & 0.8544 \\ 
% \hline
% \pname{} (Learned) & 0.6602 & 0.5340 \\
% \pname{} (Fixed) & 0.7087 & 0.4272 \\
% \hline
% \end{tabular}
\begin{tabular}{l|cc} 
\hline
\multicolumn{1}{c|}{\multirow{1}{*}{Method}} & Accuracy$\uparrow$ & MAE$\downarrow$ \\ 
\hline\hline
SimCLR \cite{simclr}        & 0.4661                                     & 1.0583                                  \\
SupCon \cite{supcon}        & 0.4660                                     & 0.9029                                  \\
DINO   \cite{dino}        & 0.3883                                     & 1.1262                                  \\ 
\hline
ORCNN \cite{orcnn}         & 0.3786                                     & 1.0970                                  \\
POE \cite{poe}           & 0.5240                                     & 0.7210                                  \\
GOL \cite{gol}           & 0.6117                                     & 0.6700                                  \\
MWR \cite{mwr}           & 0.4039                                     & 0.9420                                  \\
RnC \cite{rnc}           & 0.5437                                     & 0.8544                                  \\ 
\hline
\pname{} (Learned) & 0.6602 & 0.5340 \\
\pname{} (Fixed) & 0.7087 & 0.4272 \\
\hline
\end{tabular}
% }
% \caption{Comparison of test set accuracy and MAE of SupCon, RnC, and \pname{} (with learned or fixed critical margin) trained with training sets containing biases at critical boundaries.}
% \label{table:noise}
% \arrayrulecolor{black}
% \end{table}
    }
    \caption{Comparison of \pname{} (learned or fixed margin) vs. related methods on biased training sets at critical boundaries.}
    \label{table:noise}
\end{minipage}
\hfill
\begin{minipage}[t]{0.67\linewidth}
    \centering
    \resizebox{\linewidth}{!}{
    % \usepackage{graphicx}
% \usepackage{multirow}
% \begin{table*}[tb]
% \centering
% \resizebox{0.9\linewidth}{!}{%
\begin{tabular}{l|ccccc|ccccc} 
\hline
\multirow{2}{*}{\begin{tabular}[c]{@{}c@{}}At end of\\Training\end{tabular}} & \multicolumn{5}{c|}{Accuracy $\uparrow$} & \multicolumn{5}{c}{Mean Absolute Error $\downarrow$} \\ 
\cline{2-11}
\multicolumn{1}{c|}{} & Adiance & HID & KOA & IDRID & BRACS & Adiance & HID & KOA & IDRID & BRACS \\ 
\hline\hline
\multicolumn{1}{l}{~} & \multicolumn{10}{c}{Ablation study 1: With $C-1$ margins each set to 1} \\
% \multicolumn{1}{l}{~} & \multicolumn{10}{c}{\multirow{2}{*}{Ablation study 3: With $C-1$ margins each set to 1}} \\
% \multicolumn{1}{l}{~} & \multicolumn{10}{c}{~} \\
\hline
phase 2 & \textbf{0.6306} & \textbf{0.6492} & 0.6612 & 0.7184 & 0.5912 & \textbf{0.4000} & \textbf{0.5432} & 0.4257 & 0.4272 & 0.8228 \\
\hline
\multicolumn{1}{l}{~} & \multicolumn{10}{c}{Ablation study 2: With a single learnable value for $C-1$ margins} \\
% \multicolumn{1}{l}{~} & \multicolumn{10}{c}{\multirow{2}{*}{Ablation study 1: With a single learnable value for $C-1$ margins}} \\
% \multicolumn{1}{l}{~} & \multicolumn{10}{c}{~} \\
\hline
phase 1 & 0.5803 & 0.5916 & 0.6413 & 0.6893 & 0.5544 & 0.5500 & 0.6436 & 0.4644 & 0.5437 & 0.8649 \\
phase 2 & 0.5939 & 0.5944 & 0.6594 & 0.7184 & 0.5860 & 0.5100 & 0.6396 & 0.4457 & \textbf{0.4078} & 0.7860 \\  
\hline
\multicolumn{1}{l}{~} & \multicolumn{10}{c}{Ablation study 3: With $C-1$ learnable values for $C-1$ margins (\pname{})} \\
% \multicolumn{1}{l}{~} & \multicolumn{10}{c}{\multirow{2}{*}{Ablation study 2: With $C-1$ learnable values for $C-1$ margins}} \\
% \multicolumn{1}{l}{~} & \multicolumn{10}{c}{~} \\
\hline
phase 1 & 0.5852 & 0.6028 & 0.6600 & 0.7282 & 0.5930 & 0.5500 & 0.6292 & 0.4426 & \textbf{0.4078} & 0.7333 \\
phase 2 & 0.6302 & 0.6208 & \textbf{0.6673} & \textbf{0.7379} & \textbf{0.6035} & 0.4100 & 0.5534 & \textbf{0.4167} & \textbf{0.4078} & \textbf{0.7054} \\
\hline
\end{tabular}

    }
    \captionof{table}{Ablation study: 1) fixing all margins to one, 2) learning a single value for $C-1$ margins, and 3) learning $C-1$ margins (\pname{}).}
    \label{table:ablations}
\end{minipage}
\end{table*}

\subsection{Interpretability}
\label{sec:interpretability}
The multi-margin property of \pname{} offers a novel way to interpret data by the representation margins between ordinal classes. First, users can quantitatively examine the margins to infer the relative separation between ranks. 
For instance, in the IDRID dataset, the margins between consecutive grades are 0.35, 0.41, 0.23, and 0.30, where the largest margin (0.41) highlights the critical decision boundary—indicating the most challenging region for separation. 
A similar trend is observed in the Adience dataset, with margins of 0.11, 0.08, 0.20, 0.12, 0.13, 0.09, and 0.12, where the largest margin (0.20) occurs between age groups 8–13 and 15–20, again marking a critical boundary.

In addition, we further assess the interpretability of \pname{} by comparing its UMAP \cite{umap} visualizations to those of nominal classification (CE), SupCon, and RnC on the IDRID dataset. As shown in \cref{fig:emb_comp}, \pname{} effectively preserves the ordinal structure of the IDRID grades and establishes a wider margin between grades 1 and 2, the dataset’s most critical decision boundary. In contrast, RnC maintains the correct grade order but lacks inter-class separability. CE and SupCon form distinct clusters but fail to capture the ordinal relationships, showing the advantage of \pname{} in maintaining order while improving class separation.

\subsection{Ablation Studies}
We experiment with three variations of CLOC, by manipulating margin parameters in different ways: 1) setting a fixed value of 1 to all margins, 2) learning a single value for all $C-1$ margins, and 3) learning every $C-1$ margin.
All experiments follow the same setup described in \cref{sec:ex_setting}.

Table \ref{table:ablations} reports results for these ablation studies.
It is evident that learning all $C-1$ margins results in higher accuracy and lower MAE compared to learning a single value for all margins.
When all margins are set to 1, two out of the five datasets perform better, with a significant improvement only in the HID dataset.
Overall, the results indicate that learnable multiple margins improves performance, indicating that fixed, uniform margins between consecutive classes may not effectively capture the nuanced differences among them. This finding highlights the advantage of learnable, class-specific margins to better model these distinctions.

We also observe that introducing a second training phase focused on learning representation using margins learned in phase one consistently improves accuracy compared to jointly optimising representation and margins.

Figure \ref{fig:precautions} shows margin learning in Phase One on IDRID with and without precautions introduced in Sec. \ref{sec:training}, demonstrating that with ``precautionary'' measures, training can continue for more epochs without margin collapse.
Appendix \ref{sec:different_backbones} shows \pname{}'s accuracy increases as the number of parameters in different base models increases.

\section{Discussion}
\noindent\textbf{Beyond overall accuracy:} our result highlights a crucial insight: overall accuracy, while commonly used, may not adequately reflect the true importance of different outcomes in practical applications. 
By allowing flexible error prioritization, \pname{} proves especially valuable in contexts where critical decision thresholds influence high-stakes outcomes, such as in medical applications. 
This finding opens a promising research direction focused on developing more refined metrics for classification, tailored to reflect the varying weights of errors in real-world contexts where some mistakes carry more serious consequences than others.

\noindent\textbf{Connection between margins and decision boundaries:}
The coincidence of the largest learned margins with critical decision boundaries is intriguing (Section~\ref{sec:interpretability}). 
One possible explanation is that the initial decision guidelines were defined based on distinct visual differences, which aligned the decisions with the most prominent visual features, making it easier to achieve separation in the representation space.
Additionally, expert annotators may exercise greater caution near key boundaries, leading to more precise labeling and clearer representations.
It is important to emphasize that the margin value alone does not offer a definitive explanation, as the confounding factors like visual distinctiveness and annotator caution are difficult to disentangle. 
Instead, much like other interpretable AI methods \cite{reyes2020xai, fuhrman2022xai}, the margin value acts as an additional tool for understanding model behavior, providing valuable insights, particularly in high-stakes applications where decision boundaries are crucial.

\noindent\textbf{Limitation:} \pname{}'s performance decreases on datasets with a large number of classes, potentially due to the complexity of optimizing numerous margins (Appendix \ref{sec:many_classes}).
However, \pname{} excels in ordinal classification, especially in medical applications, which typically feature moderate class numbers and decision boundaries.

\section{Conclusion}
We introduced \pname{}, a new contrastive learning method for ordinal classification that learns an ordered representation based on the optimization of multiple margins with a novel multi-margin n-pair loss. 
\pname{} learns flexible margins that adapt to the varying importance of adjacent label errors while maintaining ordinal relationships. 
Experiments on five real-world and one synthetic dataset show that our approach outperforms others.
Beyond improved overall accuracy, \pname{} offers enhanced controllability and interpretability, enabling practitioners to prioritize error reduction between specific ranks and set decision boundaries based on application needs, making it effective for complex tasks like medical image analysis.

\section*{Acknowledgments}

G.C. acknowledges support from the Engineering and Physical Sciences Research Council (EPSRC) through grant EP/Y018036/1.

\appendix

{
    \small
    \bibliographystyle{ieeenat_fullname}
    \bibliography{ref}
}

\clearpage
\setcounter{page}{1}
\maketitlesupplementary

\section{Datasets}
\subsection{Adience dataset}
\label{app:adience}
This diverse dataset includes people from various cultures, ethnicities, backgrounds, and attire, along with image variations like additional people in the background.
The five folds has 9,268, 6,872, 5,913, 7,104 and 7,706 images respectively.
On average each fold includes 363 images in the 0-2 age group, 317 in 4-6, 317 in 8-13, 224 in 15-20, 670 in 25-32, 313 in 38-43, 114 in 48-53 and 116 in 60+ age group.
Figure \ref{fig:teaser} presents sample images from the dataset.

\subsection{Historical Colour Image Dating (HID) dataset}
\label{app:hid}
\begin{figure*}
\includegraphics[width=1\linewidth]{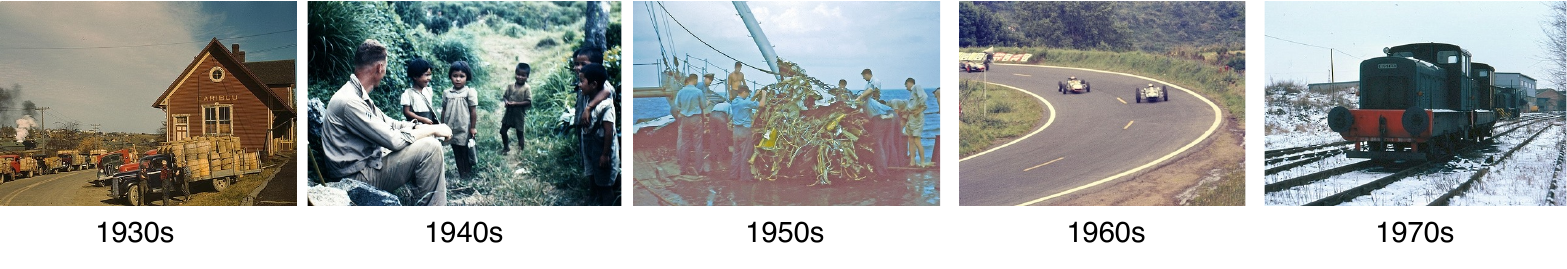}
\captionof{figure}{Examples from Historical Colour Image Dating (HID) dataset}
\label{fig:hic_examples}
\end{figure*}

This dataset features vehicles, scenery, roads, landscapes covering various seasons and contexts, occasionally including people. Figure \ref{fig:hic_examples} provides sample images from the dataset.

\subsection{Knee Osteoarthritis (KOA) dataset}
\label{app:koa}
\begin{figure*}
\includegraphics[width=\linewidth]{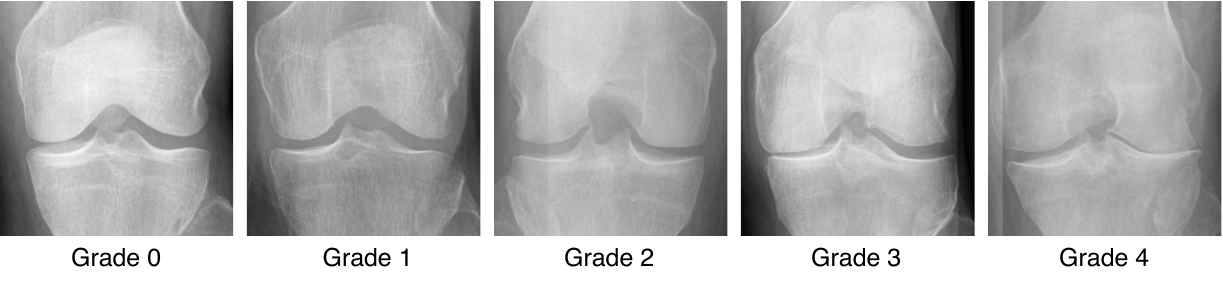}
\captionof{figure}{Examples from Knee Osteoarthritis (KOA) dataset}
\label{fig:koa_examples}    
\end{figure*}

The dataset includes 2,286 grade 0, 1,046 grade 1, 1,516 grade 2, 757 grade 3, and 173 grade 4 knee joints in the training set. The validation set contains 328 grade 0, 153 grade 1,212 grade 2, 106 grade 3, and 27 grade 4 samples. We combine these into a single training set. We evaluate using the provided testing split having 639 knee joints of grade 0, 296 of grade 1, 447 of grade 2, 223 of grade 3, and 51 of grade 4 and train using the rest of the dataset. Figure \ref{fig:koa_examples} shows sample images from the dataset.

\subsection{Indian Diabetic Retinopathy Image Dataset (IDRID) dataset}
\label{app:idrid}
The training set features 134 fundus images from grade 0, 20 from 1, 136 from 2, 74 from 3 and 49 from 4. Following a similar distribution, the testing set includes 34 fundus images from grade 0, 5 from grade 1, 32 from grade 2, 19 from grade 3 and 13 grade 4. The figure \ref{fig:teaser} shows samples from the dataset.

\subsection{BReAst Carcinoma Subtyping dataset (BRACS) dataset}
\label{app:bracs}
\begin{figure*}[!h]
\includegraphics[width=\linewidth]{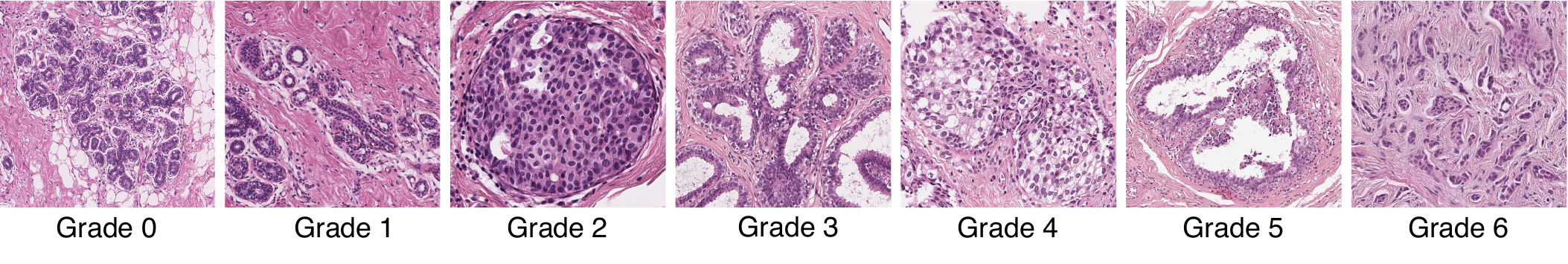}
\captionof{figure}{Examples BRACS dataset}
\label{fig:bracs_examples}
\end{figure*}

The training set has 403 images from grade 0, 757 from grade 1, 435 from grade 2, 673 from grade 3, 428 from grade 4, 705 from grade 5 and 568 from grade 6.
The esting set includes 81 images from grade 0, 79 from grade 1, 82 from grade2, 83 from grade 3, 79 from grade 4, 85 from grade 5 and 81 from grade 6. The figure \ref{fig:bracs_examples} shows samples from the dataset.

\section{Phase One Training Remarks}
\label{sec:properties}
This section outlines some remarks from the training phase one that could lead to trivial solutions, and hence should be avoided.

\textbf{Remark 1} (Margin-collapsed solution). 
Assuming that the set of all-zero margins is represented by $m_\mathcal{H} = 0$.
Although, the training can converge with all margins being $0$, a meaningful rank representation will not be learned.

\textit{Reason:} For every configuration $(\phi^*,\gamma^*,m^*_{\mathcal{H}})$, the objective function \ref{eq:overall_op} relies on the learned margins. However, when all margins are zero, the contrastive force applies to all negative pairs becomes uniform, regardless of rank differences.
Despite this, the model still converges due to the cross-entropy loss, which optimizes for standard classification rather than ordinal classification. The optimal margins remain trivial at $m^*_\mathcal{H} = 0$, because equation \ref{eq:overall_op} is minimized. Thus, $m_\mathcal{H}^*=0$ is a technically optimal since the objective function is minimized, but a degenerate solution that fails to capture ordinal relationships. $\hfill\square$

To prevent this, we initialize the margins randomly from a uniform distribution in the range $[0.5, 1.0)$, instead of initializing them close to $0$ when training. In addition, we take precautions in remark 2 to delay converging to zero.

\textbf{Remark 2} (Non-smooth activation functions). 
Jointly optimising features and margins could lead to margin collapse if the activation function used for margin parameters is non smooth around its lower bound,  that is $\leq 0$. 

\textit{Reason:} 
Suppose the activation function for margin parameters $\varphi(\mathcal{H})$, is non-smooth (i.e. non-differentiable) at lower bound and lower bound is at or below zero (like ReLU).
In the first training phase, when training accuracy improves, the $\ell_{CE}$ in the equation \ref{eq:overall_op} is effectively minimised since it directly contributes to classification.
As training further continues, to reduce the overall loss, the optimization objective then shifts to minimize \loss{}, which depends on the margins. Further, minimizing \loss{} drives $\varphi(\mathcal{H})$ towards its lower bound, hence collapsing margins to zero. $\hfill\square$

Remark 2 simply says that if the model is trained for a large number of epochs in phase one and used a non-smooth activation function that is bounded at or below 0 for margin learning parametrs, eventually the model can approach to a all-zero margin solution (remark 1) and mimic a standard classification during training phase one.

To prevent this, we employ two strategies. 
First, we avoid using activation functions with sharp, non-smooth (i.e. non-differentiable) transitions at lower bound, such as ReLU, on margin learning parameters. Instead, we use smoother activation functions like Softplus, which provide smoother gradients and approach 0 gradually. This helps avoid dead neurons and, consequently, margin collapse.
The second strategy involves training the model in two phases. In the first phase, we optimize the model for both feature learning and margin learning objectives, with early stopping. In the second stage, we re-train the model for the feature learning objective, keeping the margins frozen at the values learned during the first stage.

\section{Additional experiments}
\subsection{With A Large Number Of Classes}
\label{sec:many_classes}
We evaluate CLOC with a VGG16 \cite{vgg16} backbone on the CLAP2015 dataset \cite{clap15} in \cref{rebuttal:abl_binsize}.
CLAP2015 has 79 classes where unique ages are treated as classes.
We first align the faces using MTCNN \cite{mtcnn} following the method in MWR \cite{mwr}.
When treating each age as a separate class (79 classes), CLOC underperformed MWR (6.09 to 2.77 testset MAE). 
However, binning consecutive ages (sizes 2, 3, 4, 5) yielded competitive results for bin sizes of 3 and above. 
For fairness, we re-ran related methods using the same labels for bin size of 3, with results in \cref{rebuttal:comp_bin3}, where we can see better performance compared to the related methods.

\begin{table}[t]
    \begin{minipage}[t]{0.45\linewidth}
    \centering
    \resizebox{0.8\linewidth}{!}{%
    \begin{tabular}{lc} 
    \hline
    \begin{tabular}[c]{@{}l@{}}\# Classes \\ (Bin size)\end{tabular} & \begin{tabular}[c]{@{}c@{}}Test set\\MAE $\downarrow$\end{tabular}  \\ 
    \hline\hline
    79(1)                                               & 6.094     \\
    42(2)                                                     & 3.134     \\
    28(3)                                                     & 2.018     \\
    22(4)                                                     & 1.471     \\
    18(5)                                                     & 1.121     \\
    \hline
    \end{tabular}
    }    
    \captionof{table}{Evaluation of CLOC on CLAP2015 with different class numbers.}
    \label{rebuttal:abl_binsize}
    \end{minipage}%
    \hspace{5pt}
    \begin{minipage}[t]{0.48\linewidth}
    \centering
    \resizebox{1.0\linewidth}{!}{%
    \begin{tabular}{lc} 
    \hline
    Method      & \begin{tabular}[c]{@{}c@{}}Test set\\MAE $\downarrow$\end{tabular}  \\ 
    \hline\hline
    POE \cite{poe}       & 2.41      \\
    GOL \cite{gol}       & 2.33      \\
    MWR \cite{mwr}       & 2.25      \\ 
    \hline
    \pname{} (Ours) & 2.02      \\
    \hline
    \end{tabular}
    }
    \captionof{table}{Comparison with related methods using 28 classes in CLAP2015 (bin size=3).}
    \label{rebuttal:comp_bin3}
    \end{minipage}%
\end{table}

\subsection{Training Time Comparison}
\label{sec:runtime_comp}
 In \cref{rebuttal:runtime}, we compare runtime (in hours) for 100 epochs on the IDRID dataset using ResNet50 backbone model on a RTX 4090. CLOC phase one and two take 0.65h and 0.54h, respectively, totaling 1.19h.
 
\begin{table}[tb]
 \begin{minipage}[!t]{\linewidth}
    \centering
    \resizebox{0.35\linewidth}{!}{%
    \begin{tabular}{lc}
    \hline
    \multicolumn{1}{c}{Method} & \begin{tabular}[c]{@{}c@{}}Time\\(hours h)\end{tabular}  \\ 
    \hline
    % \multicolumn{2}{c}{Contrastive Methods}    \\ 
    \hline
    SimCLR                     & 1.21 h        \\
    SupCon                     & 0.89 h        \\
    DINO                       & 1.01 h        \\ 
    % \hline
    % \multicolumn{2}{c}{Ordinal Methods}        \\ 
    \hline
    ORCNN                      & 3.28 h        \\
    POE                        & 0.97 h        \\
    GOL                        & 1.25 h        \\
    MWR                        & 2.22 h        \\
    RnC                        & 0.51 h        \\
    \hline
    CLOC                & 1.19 h        \\
    \hline
    \end{tabular}
    }
    \captionof{table}{Training time comparison in hours.}
    \label{rebuttal:runtime}
    \end{minipage}
\end{table}

\subsection{With Different Backbone Models}
\label{sec:different_backbones}
The table \ref{rebuttal:backbones} compares \pname{}'s performance with different backbone models on IDRID dataset, where we can see a steady increase in accuracy as the size of the model (measured by number of parameters) increases.

\begin{table}[h]
\centering
\resizebox{0.6\linewidth}{!}{%
\begin{tabular}{lcc} 
\hline
Backbone    & Accuracy $\uparrow$ & MAE $\downarrow$  \\ 
\hline\hline
\begin{tabular}[c]{@{}l@{}}DenseNet121\end{tabular} & 0.6990   & 0.4854  \\
ResNet50    & 0.7379   & 0.4078  \\
VGG16       & 0.7476   & 0.4351  \\
\hline
\end{tabular}
}
\captionof{table}{Ablation study with different backbone models, arranged in the increasing number of parameters DenseNet121 $<$ ResNet50 $<$ VGG16.}
\label{rebuttal:backbones}
\end{table}

\subsection{More Visualizations}
\label{sec:more_visuals}
\begin{figure}[h]
\centering
\includegraphics[width=\linewidth]{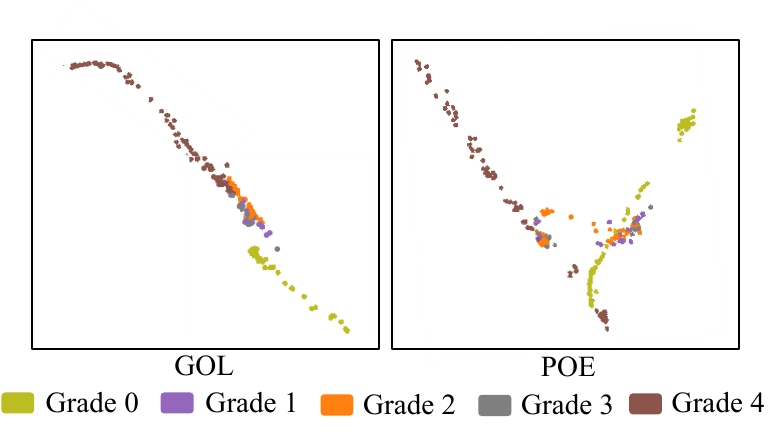}
\caption{UMAP visualizations of learned representations by GOL \cite{gol} and POE \cite{poe} for the IDRID dataset that focuses on cancer grade classification.}
\label{fig:additional_emb}
\end{figure}

The Figure \ref{fig:additional_emb} visualizes learned representation by GOL and POE by extending the Figure \ref{fig:emb_comp} in the main paper.

\end{document}